\documentclass{article}

\usepackage{microtype}
\usepackage{graphicx}
\graphicspath{{figures/}}
\usepackage{subfigure}
\usepackage{booktabs} 

\usepackage[accepted]{icml2018}

\usepackage{amsmath}
\usepackage{amssymb}
\usepackage{natbib}
\usepackage{bm}
\usepackage{mathtools}
\usepackage{dsfont}
\usepackage{wrapfig}
\usepackage{booktabs}
\usepackage{appendix}
\DeclareMathOperator{\E}{\mathbb{E}}
\DeclareMathOperator*{\argmin}{argmin}
\DeclarePairedDelimiterX{\KLdivx}[2]{\big(}{\big)}{%
  #1\;\delimsize\|\;#2%
}
\newcommand{\KLdiv}{KL\KLdivx}

\newcommand{\scare}[1]{`#1'}

 \allowdisplaybreaks
\usepackage[small,compact]{titlesec}
\titlespacing{\section}{0pt}{1ex}{0.8ex}
\titlespacing{\subsection}{0pt}{0.5ex}{0ex}
\titlespacing{\subsubsection}{0pt}{0.2ex}{0ex}
\expandafter\def\expandafter\normalsize\expandafter{%
    \normalsize
    \setlength\abovedisplayskip{2pt}
    \setlength\belowdisplayskip{2pt}
    \setlength\abovedisplayshortskip{0pt}
    \setlength\belowdisplayshortskip{0pt}
}

\begin{document}

\twocolumn[
\icmltitle{Differentially Private Continual Learning}




\begin{icmlauthorlist}
\icmlauthor{Sebastian Farquhar}{ox}
\icmlauthor{Yarin Gal}{ox}
\end{icmlauthorlist}

\icmlaffiliation{ox}{Department of Computer Science, University of Oxford, United Kingdom. Presented at the \textit{Privacy in Machine Learning and AI Workshop} at ICML 2018, Stockholm}

\icmlcorrespondingauthor{Sebastian Farquhar}{sebastian.farquhar@balliol.ox.ac.uk}

\icmlkeywords{Machine Learning, ICML, Continual Learning, Differential Privacy, GAN}

\vskip 0.3in
]



\printAffiliationsAndNotice{}  

\begin{abstract}
Catastrophic forgetting can be a significant problem for institutions that must delete historic data for privacy reasons. For example, hospitals might not be able to retain patient data permanently. But neural networks trained on recent data alone will tend to forget lessons learned on old data. We present a differentially private continual learning framework based on variational inference. We estimate the likelihood of past data given the current model using differentially private generative models of old datasets.
\end{abstract}

\section{Introduction}
Certain applications require \textit{continual} learning---splitting training into a series of tasks and discarding the data after training each task. This is often due to policy decisions: if data relate to individuals it may be unethical or illegal to keep old datasets. However, neural networks that learn from a series of datasets tend to forget lessons learned from the first datasets---a problem known as catastrophic forgetting. Institutions therefore are faced with the challenge of training models that remember relevant lessons from old datasets without storing any sensitive personal data from those datasets. Differentially Private Variational Generative Experience Replay (DP-VGER) is a dual-memory learning system that attempts to resolve this problem using differentially private generative models of old data.

\section{Related Work}
\paragraph{Continual Learning}{Several promising continual learning architectures effectively set the weights from old tasks as a prior for future training, including Elastic Weight Consolidation (EWC) \citep{Kirkpatrick2017}, Synaptic Intelligence (SI) \citep{Zenke2017}, and Variational Continual Learning (VCL) \citep{Nguyen2018}. These \scare{prior-focused} approaches struggle when presented with a long series of tasks which require the parameters to change significantly between tasks. Other approaches involve a dynamic architecture, which changes significantly as new tasks are added \citep{Razavian2014, Yosinski2014a, Donahue2014, Jung2016a, Rusu2016, Li2017}. A more general approach, which we adopt, can be thought of as \scare{likelihood-focused}---we use a generative model to estimate the likelihood of discarded datasets given the current model. This can be thought of as part of the dual-memory learning system family of approaches originating from \citep{Robins1995}. \citet{Shin2017} can be seen as a non-variational member of this family, using a recursive student-teacher framework to label generative models trained on past datasets.}

\paragraph{Differential Privacy}{Differential privacy offers a tool for quantifying the extent to which a model might reveal information about any one individual's contribution to a dataset. Without such guarantees, model inversion attacks against GANs are possible \citep{Fredrikson2015}. More formally an algorithm $\mathcal{A}$ satisfies $(\epsilon, \delta)$-differential privacy if for any two datasets $\mathcal{D}$ and $\mathcal{D}'$ which differ by a single example, for any set of outputs $\mathit{S}$:
\begin{equation}
\mathrm{Pr}[\mathcal{A}(\mathcal{D})\in \mathit{S}] \leq e^\epsilon\mathrm{Pr}[\mathcal{A}(\mathcal{D}')\in \mathit{S}] + \delta
\end{equation}

Implementations of differentially private training regimes for neural networks include works by \citet{Abadi2016} and \citet{Papernot2017}. Other authors have considered the question of training differentially private GANs \citep{Beaulieu-Jones2017, Zhang2018} broadly following \citet{Abadi2016} in the use of Gaussian noise added to clipped gradients while training the discriminator. We make use of the dp-GAN framework developed by \citet{Zhang2018}, with some modifications, to produce our generative models.}

\section{Model architecture}
Existing Variational Inference (VI) derivations for continual learning use the posterior at the end of training on one task as the prior for beginning the next \citep{Nguyen2018}. In addition to hurting multi-task performance in certain settings, this means that privacy costs accumulate over the whole of the training process. We approach the derivation of continual learning in VI differently. Instead of changing priors between tasks, we adapt the log-likelihood component of the loss to depend on past datasets. We estimate the log-likelihood component using Monte Carlo sampling from a generative model of past data. It is this generative model which we then train with attention to differential privacy.

More formally, in the continual learning context, the data are split into separate tasks $\mathcal{D}_{t} = (\mathrm{X}_t, \mathrm{Y}_t)$ whose individual examples $\mathbf{x}_{t}^{(n_{t})}$ and $y_{t}^{(n_{t})}$ are assumed to be i.i.d. Under the standard VI approach \citep{Jordan1999}, we want to find the posterior over the parameters, $p(\omega|\mathcal{D}_{0:T})$ which let us estimate a probability distribution for $y$. The posterior is generally intractable, so we introduce a tractable approximating distribution $q_{\theta}(\omega)$ and minimize the Kullback-Leibler divergence between the approximating distribution and posterior which, after applying Bayes' theorem and removing constant terms, leads to:
\begin{equation}\label{VI}
\begin{split}
q^{*}_{\theta} = \argmin_{q_{\theta} \in \mathcal{Q}} \Bigl( \KLdiv{q_{\theta}(\omega)}{p(\omega)} - \\ \E_{\omega \sim q_{\theta}(\omega)}\Bigl[\mathrm{log}p(\mathrm{Y}_{0:T}|\omega, \mathrm{X}_{0:T}) \Bigr]\Bigr).
\end{split}
\end{equation}

This is equivalent to minimizing the variational free energy of the model. The first term can be interpreted as a prior about the distributions of the parameters. The second term is data-dependent and is the negative log-likelihood of the observed data given the parameters of the model. From \ref{VI} and decomposing into a sum of separate terms for each dataset we find an expression for the variational free energy $\mathcal{F}_T$ that can be used to train on dataset $\mathcal{D}_T$:
\begin{equation}\label{eq:1}
\mathcal{F}_T \propto  \sum^{T}_{t=1}\Big( \frac{1}{T}\KLdiv{q_{\theta_{T}}(\omega)}{p(\omega)}-\mathcal{L}_{t}\Big)
\end{equation}

where $\mathcal{L}_t$ is the expected log-likelihood over $\mathcal{D}_t$:
\begin{equation}\label{eq:2}
\mathcal{L}_{t} = \sum^{N_{t}}_{n=1}\E_{\omega \sim q_{\theta_{T}}(\omega)}\left[\mathrm{log}p(\mathit{y}_{t}^{(n)}|\omega, \mathbf{x}_{t}^{(n)} \right].
\end{equation}

In order to estimate $\mathcal{F}_T$ despite not having access to old datasets, we approximate the sum in \eqref{eq:2} as an integral over a data distribution $p_t(\mathbf{x}, y)$ using a generative model.  we train a GAN $q_t(\mathbf{x}, y)$ \citep{Goodfellow2014} to produce $(\mathbf{\hat{x}}, \hat{y})$ to approximate the distribution of past datasets $p_t(\mathbf{x}, y)$ for each class in each dataset as it arrives. After that dataset is used, the data are discarded and the generator is kept.

Data from the distributions $\mathbf{\hat{x}}, \hat{y} \sim q_{1:T-1}(\mathbf{x}, y)$ can supplement the actual data for task T to create $(\mathbf{\tilde{x}}, \tilde{y}) = (\mathbf{\hat{x}} \cup \mathbf{x}, \hat{y} \cup y)$. Because we were able to separate \eqref{eq:1} as an sum of task-specific terms we can minimize it by minimizing:
\begin{align}\label{eq:4}
\mathcal{\tilde{F}} &= \E_{\omega \sim q_{\theta_{T}}(\omega)}\bigl[\mathrm{log}p(\mathit{\tilde{y}}|\omega, \mathbf{\tilde{x}}) \bigr] 
- \frac{1}{NT} \KLdiv{q_{\theta_{T}}(\omega)}{p(\omega)} 
\end{align}
which we approximate by averaging sampled mini-batches.

\subsection{Differentially Private GAN}
We implement the GAN used to estimate $\mathcal{\tilde{F}}$ in a way that allows us to establish differential privacy bounds on the generative models used. We use the \scare{dp-GAN} framework from \citet{Zhang2018} which works by clipping the gradients of each parameter and then adding Gaussian noise to the gradients at a similar scale to the gradients. By adjusting the clipping bounds and the variance of the Gaussian noise, it is possible to adjust $\epsilon$ and $\delta$ values representing the differential privacy cost of the training. \citet{Zhang2018} allow themselves some of the dataset as \scare{public} data which is used to pre-train the model and estimate gradients during training to allow grouped clipping bounds. We begin by limiting this \scare{public} data to only 1\% of the dataset rather than their 2\%. Unlike their main results, we then remove public data entirely.
\section{Experimental Results}
\begin{figure}
\hspace{-6mm}
\includegraphics[width=\columnwidth]{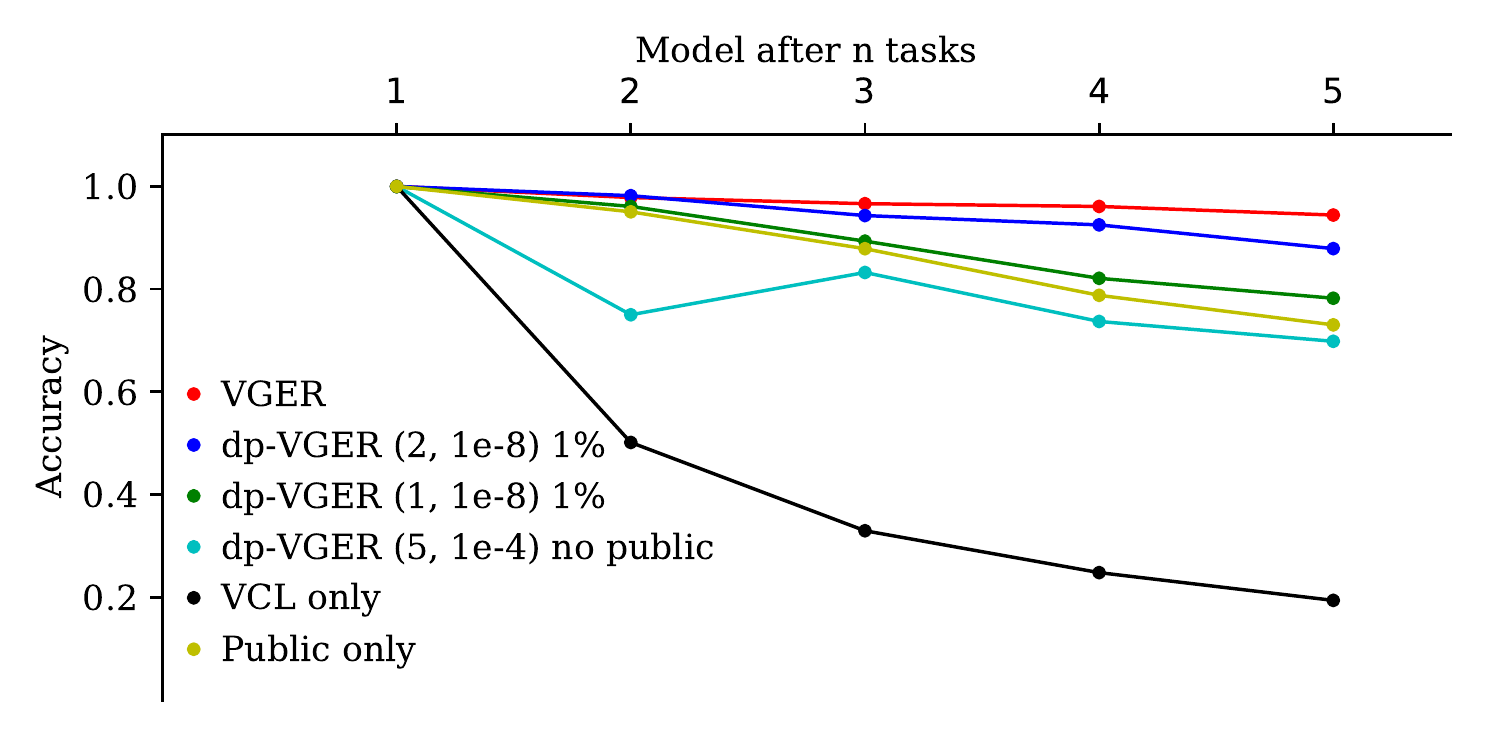}
\caption{Single-headed Split MNIST. Model accuracy is assessed on all seen tasks after training on each task. DP-VGER with public data outperforms a coreset only approach and is better when privacy bounds are looser. However, without public data performance is significantly worse.}
\label{fig:dp_single_average}
\vspace{-6mm}
\end{figure}
We test DP-VGER under three configurations against several baselines. We use a DP-VGER model with 1\% of the dataset as public data under (1, $10^{-8}$)-dp and (2, $10^{-8}$)-dp per class GAN. We also show DP-VGER with no public data under the (excessively) generous bounds (5, $10^{-4}$)-dp. We compare this against VGER with no privacy bounds, against a baseline that uses only the \scare{public} data as a density model, and against current state-of-the-art prior-focused model VCL \citep{Nguyen2018}. We evaluate using Split MNIST, first introduced by \citet{Zenke2017}.  The experiment constructs a series of five related tasks. The first task is to distinguish the digits (0, 1), then (2, 3) etc. The model is trained on each task in turn, and all old datasets are deleted. The model is then tested on all datasets including the old ones, in order to ensure that catastrophic forgetting has not occurred. Unlike \citet{Zenke2017} and \citet{Nguyen2018} we train with a single output head, rather than one head for each task.

The differentially private training of the GANs improves on simply using the public data. But without the public data, the continual learning system based on differentially private GANs significantly underperforms non-private VGER (see figure \ref{fig:dp_single_average}).

\bibliography{Mendeley}
\bibliographystyle{icml2018}
\end{document}